%% file: neurips_2019.tex
\documentclass{article}

     \PassOptionsToPackage{numbers, compress}{natbib}

    \usepackage[final]{neurips_2019}

 \usepackage{natbib}
\usepackage[utf8]{inputenc} 
\usepackage[T1]{fontenc}    
\usepackage{hyperref}       
\usepackage{url}            
\usepackage{booktabs}       
\usepackage{amsfonts}       
\usepackage{nicefrac}       
\usepackage{microtype}      
\usepackage{graphicx}
\usepackage[dvipsnames]{xcolor}
\usepackage{amssymb}
\usepackage{amsmath}
\usepackage[normalem]{ulem}
\usepackage{booktabs}
\usepackage{wrapfig}
\usepackage{subcaption}
\usepackage[font=small]{caption}

\title{Deep Crowd-Flow Prediction in Built Environments}

\author{%
  Samuel S.~Sohn \\
  Rutgers University \\
  \texttt{sss286@cs.rutgers.edu} \\
   \And
   Seonghyeon Moon \\
   Rutgers University \\
   \texttt{sm2062@cs.rutgers.edu} \\
   \And
   Honglu Zhou \\
   Rutgers University \\
   \texttt{hz289@cs.rutgers.edu} \\
   \AND
   Sejong Yoon \\
   The College of New Jersey \\
   \texttt{yoons@tcnj.edu} \\
   \And
   Vladimir Pavlovic \\
   Rutgers University \\
   \texttt{vladimir@cs.rutgers.edu} \\
   \And
   Mubbasir Kapadia \\
   Rutgers University \\
   \texttt{mk1353@cs.rutgers.edu} \\
}

\begin{document}

\maketitle

\input{Tex/0commands.tex}

\vspace{-10pt}
\begin{abstract}
\vspace{-10pt}
\input{Tex/0abstract.tex}
\end{abstract}

\vspace{-10pt}
\section{Introduction}
\vspace{-10pt}
\input{Tex/1introduction.tex}

\vspace{-10pt}
\section{Related Works}
\label{sec:related}
\vspace{-5pt}
\input{Tex/2related.tex}

\vspace{-10pt}
\section{Specification}
\vspace{-10pt}
\input{Tex/3specification.tex}

\vspace{-10pt}
\section{Synthetic Dataset Generation}
\vspace{-10pt}
\input{Tex/4dataset.tex}

\vspace{-10pt}
\section{Proposed Approach}
\vspace{-10pt}
\input{Tex/5prediction.tex}

\vspace{-10pt}
\section{Evaluation}
\vspace{-10pt}
\input{Tex/6evaluation.tex}

\vspace{-10pt}
\section{Conclusion}
\vspace{-10pt}
\input{Tex/7conclusion.tex}

\newpage
\bibliography{neurips_2019}

\newpage
\section{Appendix}
\input{Tex/8appendix.tex}

\end{document}

%% file: Tex/0commands.tex
\newcommand{\CAGE}{{\texttt{CAGE}} }
\newcommand{\SF}{social forces }

\newcommand\red[1]{\textcolor{red}{#1}}
\newcommand\orange[1]{\textcolor{orange}{#1}}
\newcommand\blue[1]{\textcolor{blue}{#1}}
\newcommand\honglu[1]{\textcolor{purple}{[HZ]~#1}}
\newcommand\mk[1]{\textcolor{purple}{[MK]~#1}}
\newcommand\sss[1]{\textcolor{darkgreen}{[SSS]~#1}}
\newcommand\sy[1]{\textcolor{royalblue}{[SY]~#1}}
\newcommand\vp[1]{\textcolor{red}{[VP]~#1}}

\newcommand\start{\textcolor{green}{<<<<<<<<START}}
\newcommand\finish{\textcolor{green}{END>>>>>>>>}}

\definecolor{darkgreen}{rgb}{0,0.5,0}
\definecolor{purple}{rgb}{0.5,0,0.5}
\definecolor{royalblue}{rgb}{0.25,0.4,1}
\newcommand\green[1]{\textcolor{darkgreen}{#1}}

\newcommand{\Env}{\mathbf{E}}
\newcommand{\Age}{\mathbf{A}}
\newcommand{\Goa}{\mathbf{G}}
\newcommand{\Capa}{\mathbf{C}}
\newcommand{\Flo}{\mathbf{F}}
\newcommand{\tM}{\mathbf{t}^M}
\newcommand{\tO}{\mathbf{t}^O}
\newcommand{\Cx}{\mathbf{C}^\mathbf{x}}
\newcommand{\Cy}{\mathbf{C}^\mathbf{y}}
\newcommand{\rx}{\hat{R}^\mathbf{x}}
\newcommand{\ry}{\hat{R}^\mathbf{y}}
\newcommand{\D}{\mathcal{D}}
\newcommand{\X}{\mathbf{X}}
\newcommand{\XO}{\mathbf{X}^O}
\newcommand{\Y}{\mathbf{Y}}
\newcommand{\x}{\mathbf{x}}
\newcommand{\y}{\mathbf{y}}
\renewcommand{\b}{\mathbf{b}}
\newcommand{\V}{\mathcal{V}}
\newcommand{\N}{\mathcal{N}}
\newcommand{\R}{\hat{R}}
\newcommand{\ModelA}{Mono-SegNet }
\newcommand{\ModelB}{Dual-SegNet }
\newcommand{\KLD}{D_{KL}}

\newcommand*{\affaddr}[1]{#1} 
\newcommand*{\affmark}[1][*]{\textsuperscript{#1}}
\newcommand*{\email}[1]{\texttt{#1}}

%% file: Tex/0abstract.tex
Predicting the behavior of crowds in complex environments is a key requirement in a multitude of application areas, including crowd and disaster management, architectural design, and urban planning. Given a crowd's immediate state, current approaches simulate crowd movement to arrive at a future state. However, most applications require the ability to predict hundreds of possible simulation outcomes (e.g., under different environment and crowd situations) at real-time rates, for which these approaches are prohibitively expensive.


In this paper, we propose an approach to instantly predict the long-term flow of crowds in arbitrarily large, realistic environments. Central to our approach is a novel  \textbf{\texttt{CAGE}} representation consisting of \textbf{C}apacity, \textbf{A}gent, \textbf{G}oal, and \textbf{E}nvironment-oriented information, which efficiently encodes and decodes crowd scenarios into compact, fixed-size representations that are environmentally lossless.
We present a framework to facilitate the accurate and efficient prediction of crowd flow in never-before-seen crowd scenarios. We conduct a series of experiments to evaluate the efficacy of our approach and showcase positive results.

%% file: Tex/1introduction.tex
Crowd flow prediction is motivated by real-world scenarios where large groups of people are put in spaces not equipped to handle their movement, leading to injuries and casualties. Notable rule-based simulators~\cite{WarpDriver} and deep sequential neural models~\cite{SocialWays,alahi2016social,PeriodicDensityPrediction,jiang2019deepurbanevent} have been proposed to mitigate the issue. Despite their strengths, they have non-negligible deficiencies, such as unsatisfactory trade-offs between computational resources, accuracy, and efficacy; impractical reliance on prior knowledge of crowd dynamics; and the innate inability to scale up to arbitrarily large environments. Inspired by these limitations, we aim to provide a practical solution that performs one-shot predictions of future crowd flow within an environment of unrestricted size under any-sized crowd.
We propose this new angle because in practice, serious crowd disasters and emergencies are often unexpected outbursts.
It is particularly important to make long-term predictions, because of their potential to aid in both timely planning and providing analyses.


Our contributions can be summarized as follows.
We propose \texttt{CAGE}, encoding and decoding techniques that enables the capability to handle arbitrarily large discrete environments and crowd situations. \texttt{CAGE} is generic and can be plugged in easily to any existing method, which we demonstrate with our predictive model.
We also formalize the task of predicting long-term crowd flows in one shot, a new direction for crowd dynamics prediction with high potential.
Finally, we present a unified framework that exploits a modified SegNet~\cite{DBLP:journals/corr/BadrinarayananK15}, a deep convolutional neural network, to predict in an efficient and accurate manner. It is used to predict for sparse and dense crowd situations, and it successfully tackles both.
The proposed approach can effectively provide thorough understanding towards the future crowd flow of an arbitrary scene, and it has the potential to further aid crowd control, behavior understanding, public security management, urban traffic, architectural design, etc.

%% file: Tex/2related.tex
\vspace{-5pt}
\label{sec:taxonomy}
Crowd flow prediction is a spatio-temporal problem; it attempts to determine the future activity of the crowd \cite{zhang2017deep,jin2018spatio,rudenko2019human}.
The problem has taken two forms: \emph{micro-predictions} on sparse crowds and \emph{macro-predictions} on the dense crowds, both of which can either be predictions in the short-term or long-term. Macro-predictions are most intuitively performed from a top-down orthogonal view of an environment; e.g., Periodic-CRN \cite{PeriodicDensityPrediction} and DeepUrbanEvent \cite{jiang2019deepurbanevent} make short-term predictions that are environment-centered. Contrarily, micro-predictions are performed over a set of focal agents. Notable examples include WarpDriver \cite{WarpDriver}, which predicts the short-term behavior of agents, Social LSTM \cite{alahi2016social} and Social Ways \cite{SocialWays}, which predict the short-term behavior of humans. 
Tackling short-term crowd flow prediction
implies the use of simulations during inference, regardless of whether by rule-based simulators or deep-learning-based models.
For long-term prediction, this becomes impractical.
In contrast to the previous works, we perform both macro- and micro-predictions for agent behaviors (and in the future, human behaviors) in the long term. 




The Convolutional Neural Network (CNN) \cite{lecun1989backpropagation} has achieved substantial success in a great number of tasks \cite{luo2018fast,huang2017densely,zhang2019graphical}. It has been prevalently deployed in the domain of crowds \cite{DBLP:journals/corr/SindagiP17,tripathi2019convolutional}. The powerful performance of CNNs were shown especially for crowd counting \cite{boominathan2016crowdnet,huang2017body} and density estimation \cite{liu2018decidenet,yogameena2017computer}. For other crowd dynamics prediction tasks, CNNs also demonstrated their effectiveness. For example, Behavior-CNN \cite{yi2016pedestrian} was proposed to model pedestrian behaviors in crowded scenes with applications to walking path prediction, destination prediction, and tracking. Variants of CNNs, especially ones combined with Recurrent Neural Networks, are commonly used to predict short-term crowd dynamics \cite{jiang2019deepurbanevent,PeriodicDensityPrediction}. Following the same trend, we favor exploiting a deep convolutional encoder-decoder architecture, but to predict long-term crowd flows.

%% file: Tex/3specification.tex


In this section, we describe our novel \CAGE encoding and decoding techniques.

\vspace{-10pt}
\subsection{Preliminaries}
\vspace{-5pt}
We formalize the problem of crowd flow prediction as follows.
As raw input, we are given a matrix of unique initial agent locations $\Age$, where agent locations are 1, and all others are zero; a corresponding matrix of shared agent goal locations $\hat{\Goa}$; and a 2D discretization of a built environment $\Env$ into a matrix.
The environment and the two location matrices have fixed dimension of $n \times n$. We use index pair $(i, j)$ to denote a location in the matrices, where $i = 1..n$ and $j = 1..n$.
The crowd flow prediction output $\Y$ is the average crowd density over an entire simulation (i.e., crowd flow) for each navigable cell in environment $\Env$.

For the agent goals, we make an assumption that each agent chooses its shared goal as the closest one. We then transform the goal locations into distance information using distance function $\D$, where $\Goa = \D(\hat{\Goa})$ converts every cell to the shortest path length from the cell to the closest goal.
The resulting $\Goa$ encodes not only where goals are located (i.e., $\Goa_{i,j} = 0$), but also how close an agent is to its goal. By performing a greedy search to minimize the value on $\Goa$, an agent is able to reach its goal. However, the implication of function $\D$ is that agents can perform pathfinding to find the shortest path to its goal, which is nearly universal in crowd simulators.

Thus far, the input encoding focused on capturing agent information. 
The scale of the environment is dependent on the diameter of agents, which is equal to the width of a grid cell. This means that a grid cell can represent an exact capacity of one agent per grid cell.
However, there is a certain inflexibility to this representation. Namely, the dimensions $n \times n$ must be able to capture the extent of both existing and future environments, making it difficult to choose an effective $n$.
We resolve this problem by altering the fixed-size agent-centric representation into a environment-centric one. We propose \CAGE, an encoding and decoding technique with the capabilities to serve any environment and crowd context with unlimited size.

\vspace{-10pt}
\subsection{\CAGE Encoding}
\begin{wrapfigure}{l}{6.5cm}
	\centering
	\vspace{-23pt}
	\includegraphics[scale=0.2]{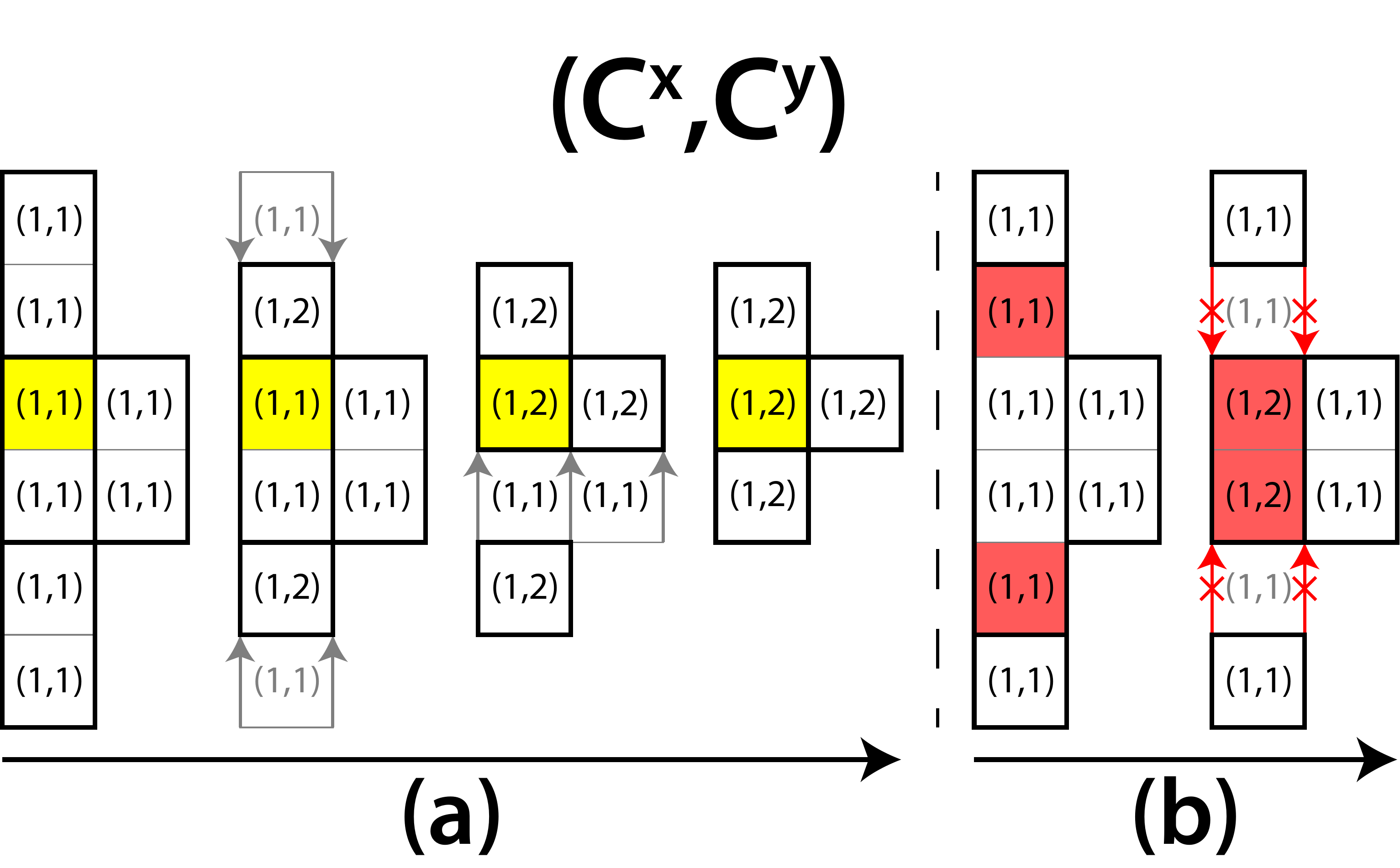}
	\caption{Shows the preservation of consistency with the manual compression of bolded regions. (a) depicts the change in $\Cx$ and $\Cy$ as cells are compressed in the direction of the arrows. (b) depicts a disruption in the environment, where compression results in two red cells becoming adjacent to cells that were not originally navigable.
	}
	\label{fig:RegionCompression}
	\vspace{-20pt}
\end{wrapfigure}

\vspace{-5pt}
Given a real environment in its discretized form $\Env$ (whose grid dimensions exceed $n \times n$), we apply a compression method that preserves environmental information. This is achieved by maintaining the local navigability (referred to as \emph{consistency}) between cells, while warping their capacities (i.e., the maximum number of agents that can occupy them).
For this, we require two additional ``channels'' of information $\Cx_{n \times n}$ and $\Cy_{n \times n}$ for storing capacities: one for capacity along the $x$-axis and the other for capacity along the $y$-axis.
Initially both $\Cx$ and $\Cy$ consists of ones, meaning that each cell has a capacity of 1 agent per cell.
By manipulating $\Cx$ and $\Cy$, regions in $\Env$ are able to be compressed, which affects channels $\Age$ and $\Goa$ as well.
Figure~\ref{fig:RegionCompression}(a) demonstrates the effect of compression on $\Cx$ and $\Cy$. At the yellow cell $(i, j)$, the capacity becomes $\big(\Cx_{i,j} \cdot \Cy_{i,j}\big)$ or 2 agents/cell, and $\Env_{i,j}$ corresponds to a $2 \times 1$ space, instead of $1 \times 1$. As long as consistency is maintained, cells can have their capacities increased to other integer values. As a result, this will compress parts of the environment.
A counter-example violating consistency is shown in Figure~\ref{fig:RegionCompression}(b). We provide a more detailed explanation of compression in Appendix~\ref{app:compression}.


The changes in $\Age$ and $\Goa$ that accompany the compression of $\Env$ are as follows. By adding $\Cx$ and $\Cy$, $\Age$ represents agent density, instead of agent locations. In Figure~\ref{fig:RegionCompression}(a), 
the compressed yellow cell's value in $\Age$ becomes the number of agents among the corresponding uncompressed cells divided by the capacity of the cell.
For $\Goa$, each of its cells takes on the minimum $\Goa$-value of the cells that are being compressed into the yellow cell, as this is consistent with the aforementioned use of $\Goa$ for greedy search.
The resulting channels constitute the \CAGE representation $\X = \big[ \Cx, \Cy, \Age, \Goa, \Env \big]$.

After compression, the original dimensions $p \times q$ remain the same as the uncompressed environment, but there is an increase in the amount of non-navigable cells. The same outer rows and columns that are non-navigable can be removed from every channel, decreasing the dimensions of the compressed representation by $(p - n)$ along the $y$-axis and $(q - n)$ along the $x$-axis.
This encoding technique is now equipped to represent an infinite number of compressed environments within $n \times n$.




\vspace{-10pt}
\subsection{\CAGE Decoding}
\vspace{-5pt}
After the crowd density $\Y$ is predicted on the compressed representation $\X$, decompression can be achieved by having temporarily stored both the original and compressed locations and sizes of each region $\R \in R$. The compressed information locates the region's data from $\Cx$, $\Cy$, and $\Y$, and the original information delimits where the data should be decompressed to.
In order to expand a region to its original size, the density predicted in each cell $(i,j)$ will be uniformly divided across $\big( \Cx_{i,j} \cdot \Cy_{i,j} \big)$-many cells, which comprised cell $(i,j)$ after compression.


%% file: Tex/4dataset.tex
The synthetic generation of crowd flow datasets is primarily motivated by the difficulty of acquiring a large amount of real data for crowd activities in varied environments. The development of an adequate generator for both built environments and crowd flow data affords flexible control over scenarios of interest, which would otherwise be infeasible with real crowds.

\vspace{-10pt}
\subsection{\CAGE Input}
\vspace{-5pt}

\begin{wrapfigure}{r}{7.2cm}
	\centering
	\vspace{-20pt}
	\includegraphics[scale=0.34]{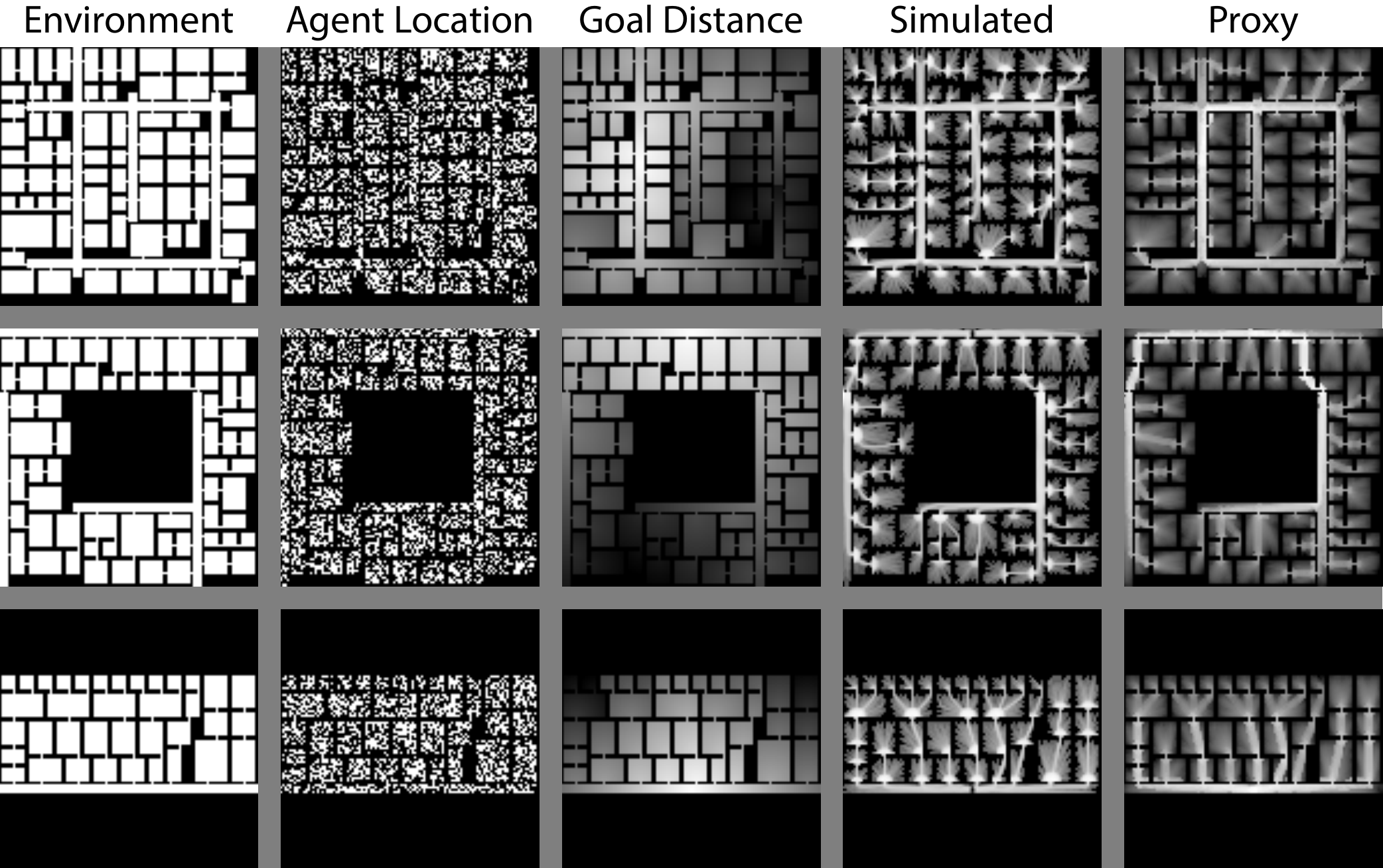}
	\caption{Shows $\Env$, $\Age$, $\Goa$, and the similarity of the proxy crowd flow and \textbf{social-force}-simulated crowd flow.}
	\label{fig:FloorplanTable}
	\vspace{-15pt}
\end{wrapfigure}

The components of the \CAGE representation define the dataset's input $X$ of initial conditions.
For the environmental component $\Env$, we utilize a multi-phase generation of floorplans for built environments with no small obstacles. This is based on a floorplan typology observed in modern architecture by Dogan et al~\cite{FloorplanTypology}, which we repurpose for floorplan generation instead of classification. The three stages of generation are (a) forming an exterior shape to the floorplan, (b) applying an interior organization of hallways, and (c) populating the floorplan with rooms interconnected toward the closest hallways.

The generated floorplans must then be accompanied by both the other input components in the \CAGE representation and the ground truth crowd flow $\hat{\Y}$.
In the dataset, the number of agents and the type of crowd flow output create four major groups, each of which have a uniform distribution of floorplan types (Figure~\ref{fig:FloorplanTable}). Namely, each group is a combination of either sparse or dense crowds and either proxy or simulated crowd flow.
The separation of sparse crowds from dense crowds is motivated by the visual differences between their respective crowd flows. When there are relatively few agents (e.g., $\leq$ 25 agents), the crowd flow is constituted of the fine paths that agents travelled along, which likens the prediction of crowd flow to finding the shortest paths from the agents to their goals (i.e., the long-term micro-predictions described in Section~\ref{sec:related}). On the contrary, when there are relatively many agents (e.g., $\geq$ 0.01 agents/m$^2$), the crowd flow becomes stigmergic, where the focus on the individual is lost when its joins the crowd (i.e., long-term macro-predictions).

\vspace{-10pt}
\subsection{Crowd Flow Output}
\vspace{-5pt}

In this work, the proxy crowd flow is realized through static generation, as opposed to dynamic generation by means of a crowd simulator for simulated crowd flow (in this case, the social force model~\cite{socialforces}). This crowd flow serves as a proxy for simulated crowd flow and makes the generation of data is made significantly faster. Compared to that of the simulated crowd flow, its visual features are less fine, which simplifies the prediction problem.
For sparse crowds, proxy crowd flow is generated by overlaying the shortest paths from individual agents to their goals (Figure~\ref{fig:VisualComparison}). For dense crowds, cohesive groups of agents have their group paths overlaid (Figure~\ref{fig:FloorplanTable}). The resulting difference between sparse crowds and dense crowds is that group paths capture the congestion caused by bottlenecks, while individual paths do not.

%% file: Tex/5prediction.tex
In accordance with the division of sparse and dense crowds, we establish the following learning goals, which apply to any potential model: ($\Goa$) to find the paths from relatively few agents to their shared goal and ($\Env$) to capture the impact of the environment on dense crowds.

\begin{wrapfigure}{l}{7.3cm}
    \centering
    \vspace{-17pt}
    \includegraphics[scale=0.25]{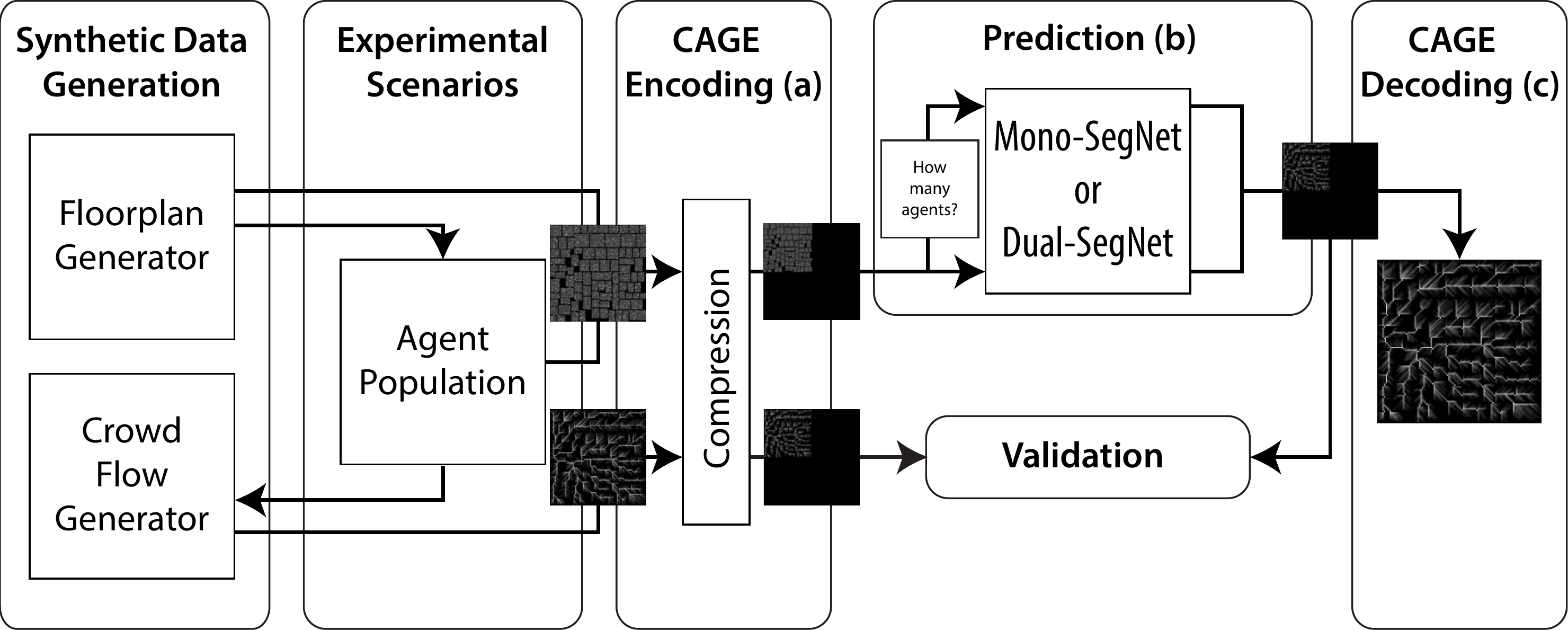}
    \caption{Shows the pipeline of the framework, which produces synthetic data based on experimental scenarios, \texttt{CAGE}-encodes it, predicts the crowd flow, and decompresses it to fit the original environment.}
	\label{fig:Framework}
    \vspace{-15pt}
\end{wrapfigure}

The proposed framework is presented in Figure~\ref{fig:Framework}. The pipeline consists of three major stages: (a) the \CAGE-encoding of raw data, (b) the prediction of crowd flow, and (c) the \CAGE-decoding of the prediction. The predictor takes as input the \CAGE representation, which is computed by compressing the original environment if necessary. If compression has occurred, then the decompression stage is necessary in order to view the crowd flow back in the original environment. Since \CAGE-encoded representations resemble images (i.e., for each channel, horizontally and vertically adjacent cell-dependencies and fixed dimensions), for the predictor (b), we exploit SegNet \cite{DBLP:journals/corr/BadrinarayananK15} to facilitate the prediction of the complex crowd flows. In particular, we substitute the last \textit{softmax} layer in SegNet with a \textit{sigmoid} layer. The resulting modified SegNet is utilized for the crowd flow prediction task on top of our proposed \CAGE representation.

To accomplish the two distinct learning goals $\Goa$ and $\Env$, we build an ensemble of models.
The baseline is the \ModelA, which is the modified SegNet that has been trained on both sparse and dense crowds.
The improved version \ModelB consists of two models: \ModelA for dense crowds and another SegNet model trained solely on sparse crowds. For the training, weights were all initialized following \cite{he2015delving}. Stochastic gradient descent with a fixed learning rate of $0.001$ and momentum of $0.9$ is used as the optimizer \cite{bottou2010large}. The batch size was set to $64$, and the training set is shuffled before each epoch. We use \textit{mean absolute error} ($MAE$) as the objective function. We trained the model with $200$ epochs and use the same set of hyper-parameters for each of the model variant.

%% file: Tex/6evaluation.tex
\subsection{Quantitative Results and Analysis}
In order to evaluate the quality of the predictions for each model on its corresponding dataset, we compare the cell-wise differences between the predicted crowd flow and the ground truth crowd flow.

The metrics included in Table~\ref{tab:QuantEval} are $MAE$ and \emph{Kullback-–Leibler divergence} ($D_{KL}$).
For the $\Env$-centric goals, low $MAE$ is indicative of good performance.
This score highlights the success of \ModelA, and thereby \ModelB, since \ModelB utilizes \ModelA for its predictions of dense crowds.
For the $\Goa$-centric goals, $MAE$ cannot be considered in the same way, because the ground truth crowd flow is comparatively sparse to that of $\Env$-centric goals. If, for instance, a predictor outputs no density, the $MAE$ for the $\Goa$-centric goal will be inherently lower than for the $\Env$-centric goal.
Therefore, $\KLD$ is needed to gain better insight into the $\Goa$-centric goals. A low $\KLD$ indicates that the predicted crowd flow is concentrated over where the ground truth crowd flow would be.
This shows that despite having a low $MAE$, \ModelA performs poorly for the $\Goa$-centric goal, because the $\KLD$ is high.
On the other hand, the $\KLD$ of \ModelB is comparatively lower, evidencing its good performance for the $\Goa$-centric goal.
\begin{wraptable}{r}{6cm}
\centering
\begin{tabular}{@{}llrrrr@{}}
\toprule
Model   & Goal & $MAE$            & $D_{KL}$     
\\ \midrule
\ModelA & $\Env$    & \emph{0.026}  & \emph{0.189}     
\\
\ModelA & $\Goa$    & 0.008  & 4.388     
\\
\ModelB & $\Env$    & 0.026  & 0.189     
\\
\ModelB & $\Goa$    & 0.005  & \emph{0.302}     
\\ \bottomrule
\end{tabular}
\caption{Shows the average performance of \ModelA and \ModelB on $\Env$- and $\Goa$-centric goals over 3,000 test cases each.}
\label{tab:QuantEval}
\vspace{-20pt}
\end{wraptable}

\begin{wrapfigure}{l}{6.5cm}
	\centering
	\vspace{-15pt}
	\includegraphics[scale=0.375]{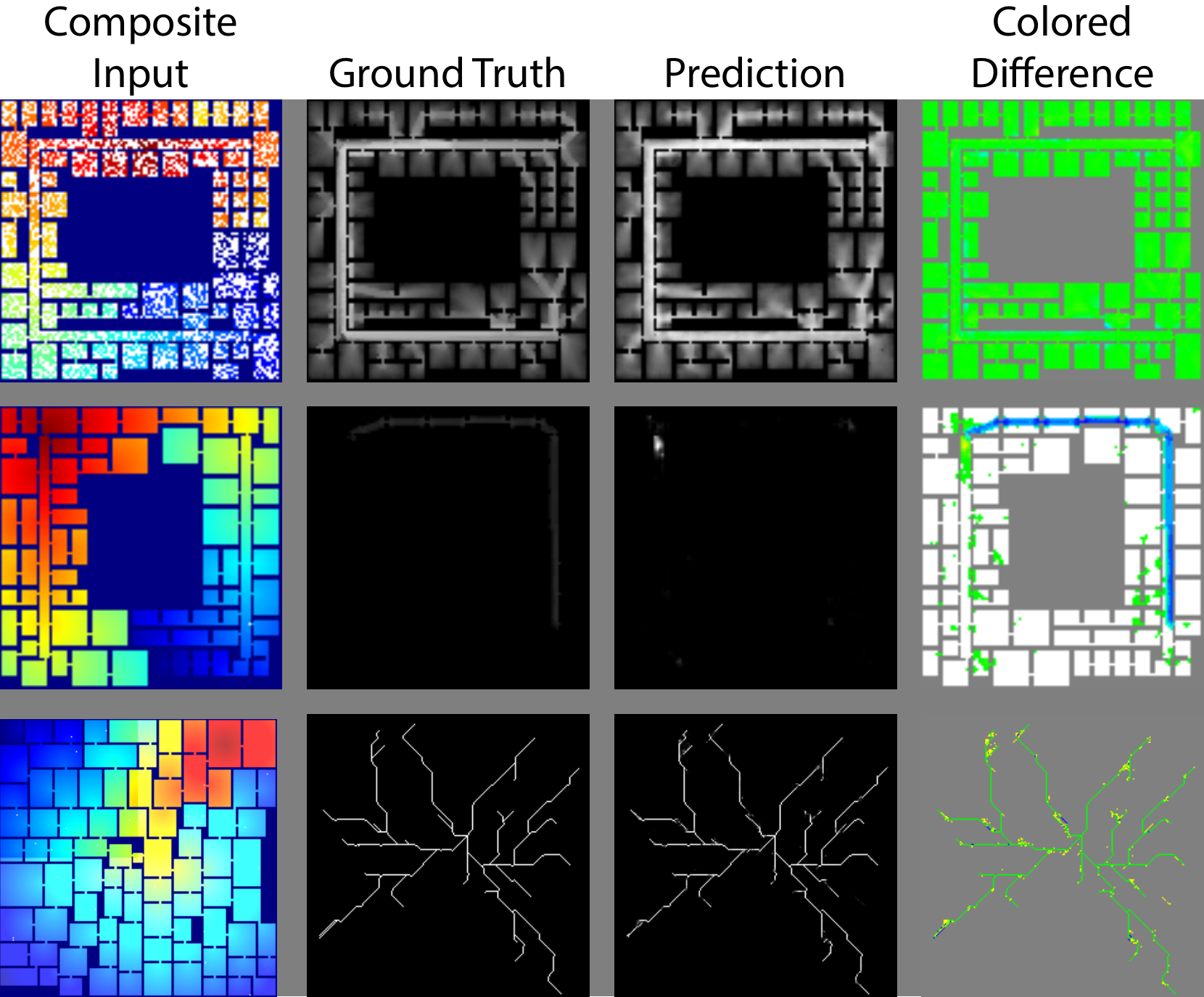}
	\caption{Shows a visual evaluation of prediction with a colored difference, which maps $\big( (\Y - \hat{\Y}) / 2 + 0.5 \big)$ to a heatmap spectrum. True positives are green, false positives tend towards red, and false negatives tend towards blue. From top to bottom, the predictions were made by \ModelA for a dense crowd, then for a sparse crowd, and \ModelB for a sparse crowd.}
	\label{fig:VisualComparison}
	\vspace{-15pt}
\end{wrapfigure}


\vspace{-15pt}
\subsection{Qualitative Results and Analysis}

The quantitative evaluation provides insights at large, but the capabilities of our models are also well-conveyed through visual comparisons between predictions and ground truth.
Figure~\ref{fig:VisualComparison} shows the results of the two models for the $\Env$- and $\Goa$-centric goals. Although there are four goals defined in Table~\ref{tab:QuantEval}, \ModelB utilizes \ModelA for its $\Env$-centric goal, so it was not given a row in Figure~\ref{fig:VisualComparison}.


The performance observed in Table~\ref{tab:QuantEval} is well-evidenced by the results in Figure~\ref{fig:VisualComparison}, which shows from left to right: the input, ground truth, prediction, and difference between the ground truth and prediction. Although the input is encoded in the \CAGE representation, its $\Age$, $\Goa$, and $\Env$ components are able to be composited. $\Goa$ appears to visually encode $\Env$, so it was used in place of $\Env$ and combined with the agent locations in $\Age$ (in white). In order to improve contrast between $\Goa$ and $\Age$, $\Goa$ was converted into a heatmap, where the red hue is closest to the goal.
The difference between the ground truth and prediction is visualized by a Colored Difference image. When the predicted crowd flow is equivalent to the ground truth crowd flow and non-zero, the hue is green. As the prediction diverges, its hue becomes either more blue (when it underpredicts the crowd flow) or more yellow (when it produces a phantom crowd flow).

From top to bottom in Figure~\ref{fig:VisualComparison}, starting with \ModelA and \ModelB's $\Env$-centric goal, the positive performance is conveyed by the abundance of green. This is can also be confirmed by viewing the ground truth and prediction images separately. Meanwhile, \ModelA's $\Goa$-centric goal was unsuccessful, as depicted by the Colored Difference, wherein the entire ground truth flow is colored a vivid blue. After having trained a separate model only for its $\Goa$-centric goal, \ModelB shows excellent performance in predicting not only the path of one agent, but also those of numerous others. These fine paths are highlighted in green, and there is very little phantom crowd flow.
Appendix~\ref{app:qualitative} and \ref{app:qualitative2} showcase numerous compelling test results for dense and sparse crowds respectively.


%% file: Tex/7conclusion.tex
We develop solutions for practical scenarios and challenging tasks where we need to predict crowd dynamics at length in real-time. We achieve this by estimating the aggregate trajectories of individual agents over an entire simulation instead of simulating each frame one at a time, and by formalizing a novel \CAGE representation. This representation is exploited by a general purpose learning framework which facilitates the prediction of crowd flow on environments of any size. The predictive component of this framework is a deep neural network model. This work focuses on tackling a simplified version of crowd flow by predicting on proxy crowd flow instead of simulated crowd flow. However, our proxy has good resemblance to simulated results, and with our application of \ModelB, we are able to achieve good performance on both predictions of dense crowds as well as sparse crowds.

%% file: Tex/8appendix.tex
\subsection{\CAGE Compression Method}
\label{app:compression}

\begin{wrapfigure}{r}{5cm}
	\centering
	\vspace{-10pt}
	\includegraphics[scale=0.15]{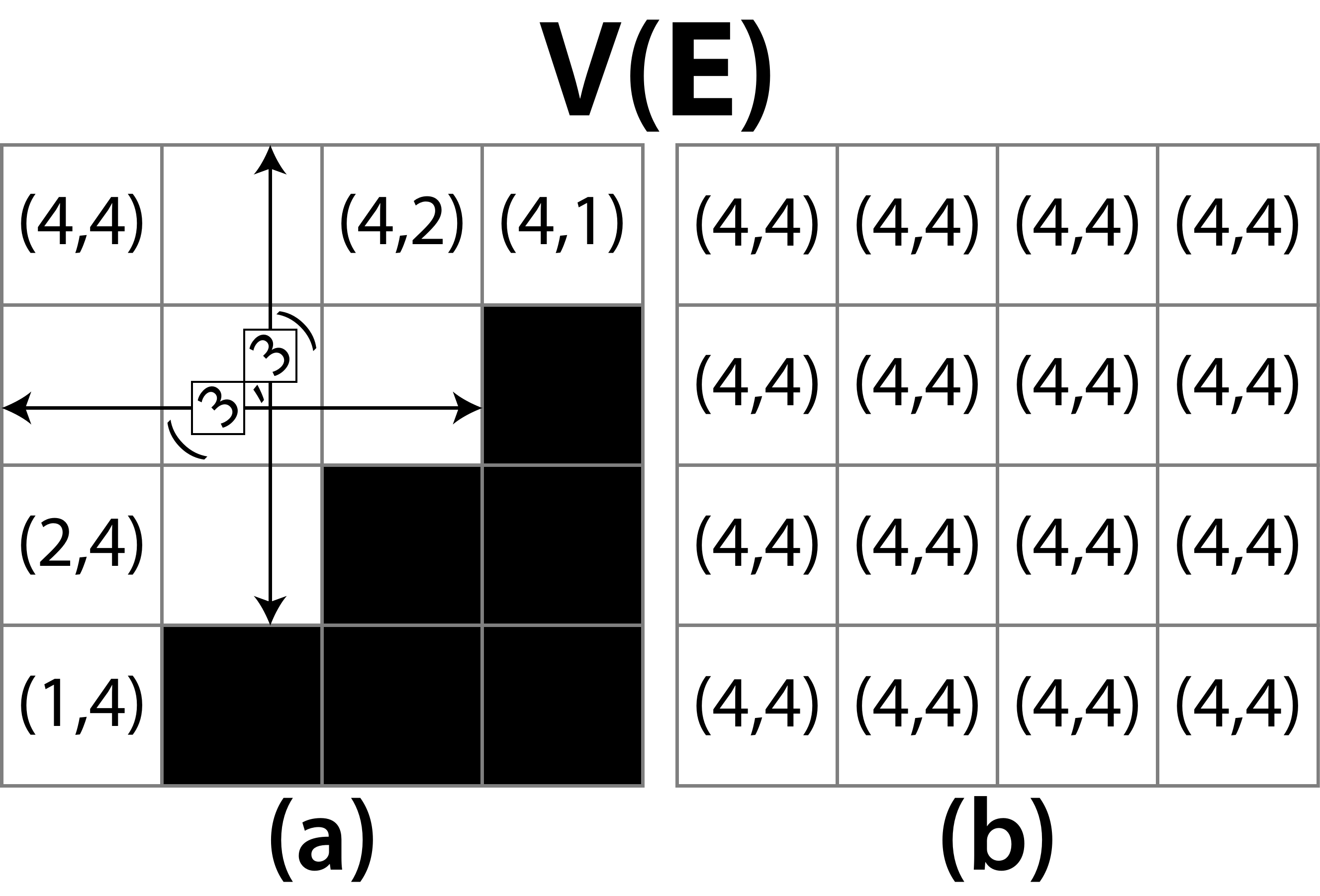}
	\caption{Matrix (a) shows an environment $\Env$ which is maximally incompressible, while (b) shows one that is maximally compressible. The cell values in both (a) and (b) have the tuple values of $\V(\Env)$, which encodes visibility. The cells indicate where $\Env$ is navigable (white) and non-navigable (black).}
	\label{fig:IncompressibleRoom}
	\vspace{-10pt}
\end{wrapfigure}

Given a real environment in its discretized form $\Env$ (whose dimensions exceed $n \times n$), let us apply a visibility function $\V$ which for each navigable cell $(i, j)$ in $\Env$, assigns a tuple
\begin{equation*}
    \big(1 + \N(i, j)_\x + \N(i, j)_{-\x},~ 1 + \N(i, j)_\y + \N(i, j)_{-\y} \big), 
\end{equation*}
where $\x = [1, 0]$, $\y = [0, 1]$ are unit vectors, and function $\N(i, j)_\b$ computes the number of navigable cells visible from $(i,j)$ along the unit vector $\b$ (see two examples in Figure~\ref{fig:IncompressibleRoom}).
Then, $\V(\Env)$ has segmented $\Env$ into a set of regions $R$ where the tuple values of each region's cells are the same. The purpose of using the visibility function is to encode navigability in the environment along the coordinate axes. Thereby, each region $\R \in R$ represents a maximal unit of space (or maximal set of neighboring cells) that is similarly navigable and therefore compressible (Figure~\ref{fig:IncompressibleRoom}).
A region's width and height are represented by $\rx$ and $\ry$ respectively, and denoted as the number of cells along each of the axis.



Two constraints are needed to ensure that compression is consistent. Namely, for any two regions $\R_i$ and $\R_j$, if they are contiguous along axis $\b$, they must (\textbf{Constraint 1}) have the same value for $\Capa^{\bot \b}$ along the orthogonal axis $\bot \b$ and (\textbf{Constraint 2}) have the same region length along axis $\bot \b$ (i.e., $\R_i^{\bot \b} = \R_j^{\bot \b}$).
These constraints ensure that neither visibility nor navigability are disrupted by compression.

\newpage
\subsection{Additional Qualitative Results for Dense Crowds}
\label{app:qualitative}

\begin{figure}[!h]
	\centering
	\includegraphics[scale=0.27]{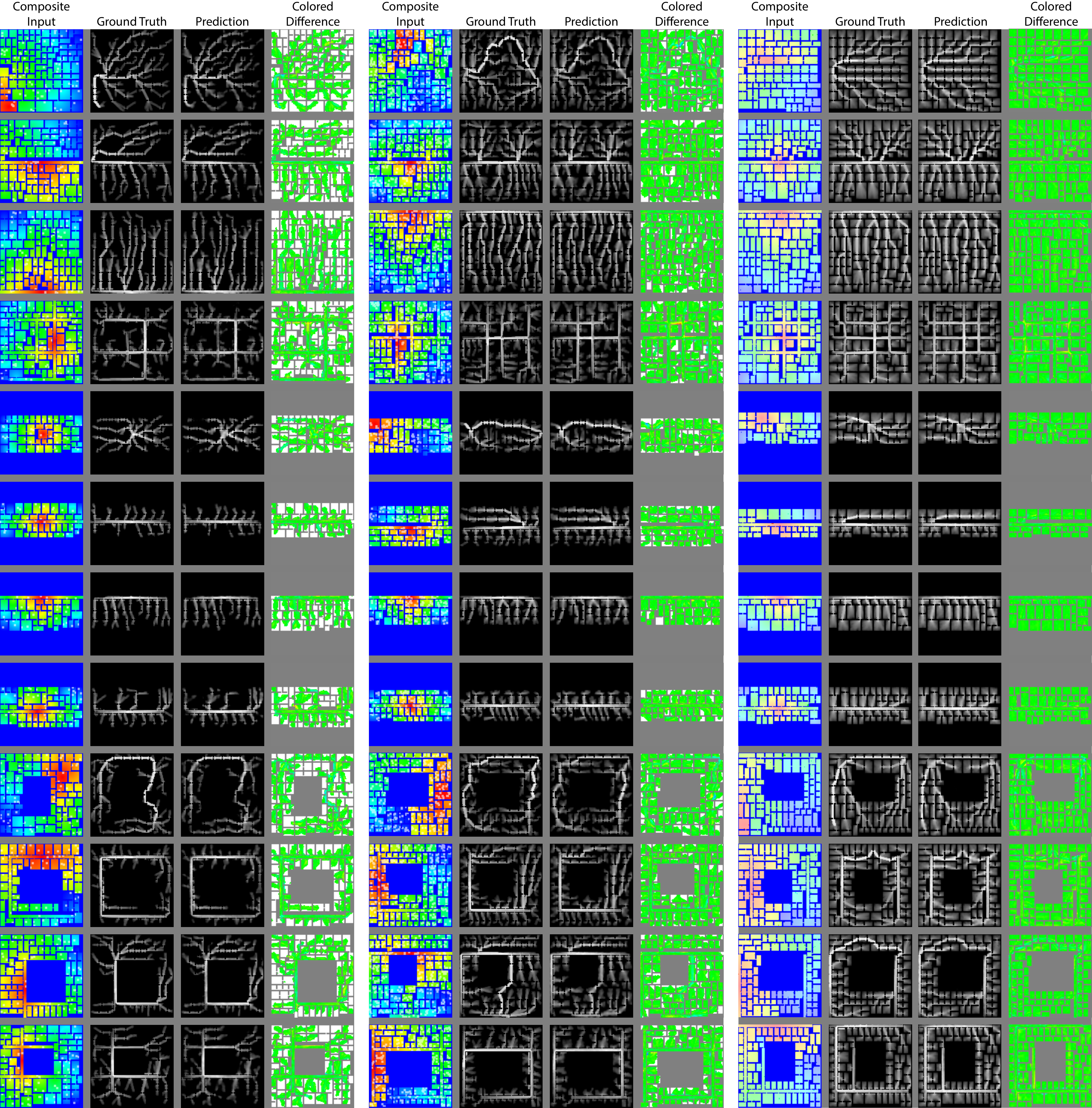}
	\caption{Shows a visual evaluation of prediction with a colored difference. True positives are green, false positives tend towards red, and false negatives tend towards blue. From left to right, the columns correspond to low, medium, and high density, respectively, but none are sparse. Each row shows a different type of built environment.}
	\label{fig:appendix}
\end{figure}

\newpage
\subsection{Additional Qualitative Results for Sparse Crowds}
\label{app:qualitative2}

\begin{figure}[!h]
	\centering
	\includegraphics[scale=0.3]{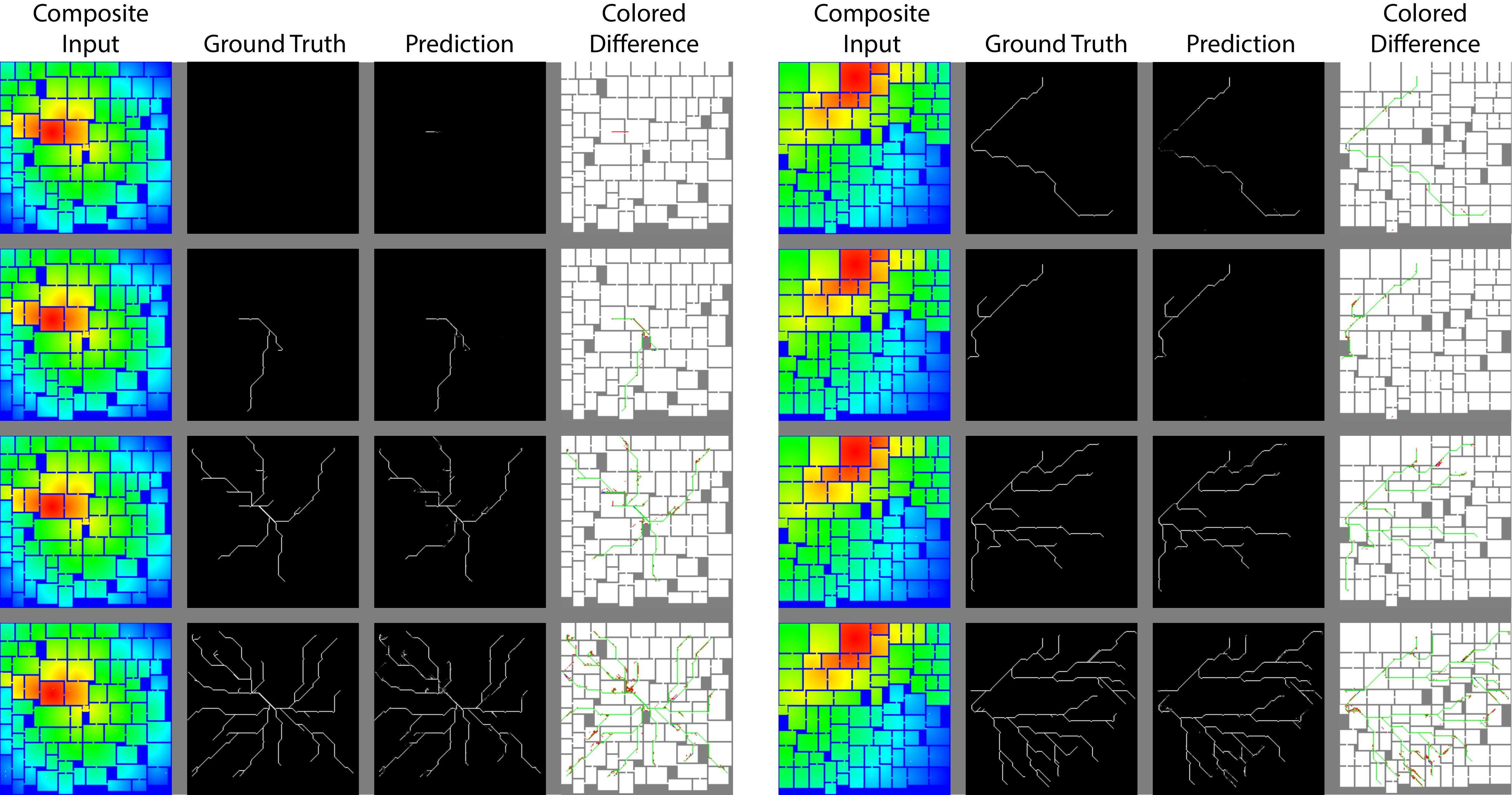}
	\caption{Shows a visual evaluation of prediction with a colored difference. True positives are green, false positives are red, and false negatives are blue. Both columns showcase prediction results on sparse crowds for one environment each. As the rows gets lower, the number of agents increases.}
	\label{fig:appendix}
\end{figure}

%% file: neurips_2019.bbl
\begin{thebibliography}{10}\itemsep=-1pt

\bibitem{alahi2016social}
A.~Alahi, K.~Goel, V.~Ramanathan, A.~Robicquet, L.~Fei-Fei, and S.~Savarese.
\newblock Social lstm: Human trajectory prediction in crowded spaces.
\newblock In {\em Proceedings of the IEEE conference on computer vision and
  pattern recognition}, pages 961--971, 2016.

\bibitem{SocialWays}
J.~Amirian, J.-B. Hayet, and J.~Pettre.
\newblock Social ways: Learning multi-modal distributions of pedestrian
  trajectories with gans.
\newblock In {\em The IEEE Conference on Computer Vision and Pattern
  Recognition (CVPR) Workshops}, June 2019.

\bibitem{DBLP:journals/corr/BadrinarayananK15}
V.~Badrinarayanan, A.~Kendall, and R.~Cipolla.
\newblock Segnet: {A} deep convolutional encoder-decoder architecture for image
  segmentation.
\newblock {\em CoRR}, abs/1511.00561, 2015.

\bibitem{boominathan2016crowdnet}
L.~Boominathan, S.~S. Kruthiventi, and R.~V. Babu.
\newblock Crowdnet: A deep convolutional network for dense crowd counting.
\newblock In {\em Proceedings of the 24th ACM international conference on
  Multimedia}, pages 640--644. ACM, 2016.

\bibitem{bottou2010large}
L.~Bottou.
\newblock Large-scale machine learning with stochastic gradient descent.
\newblock In {\em Proceedings of COMPSTAT'2010}, pages 177--186. Springer,
  2010.

\bibitem{FloorplanTypology}
T.~Dogan, E.~Saratsis, and C.~Reinhart.
\newblock The optimization potential of floor-plan typologies in early design
  energy modeling.
\newblock 12 2015.

\bibitem{he2015delving}
K.~He, X.~Zhang, S.~Ren, and J.~Sun.
\newblock Delving deep into rectifiers: Surpassing human-level performance on
  imagenet classification.
\newblock In {\em Proceedings of the IEEE international conference on computer
  vision}, pages 1026--1034, 2015.

\bibitem{socialforces}
D.~Helbing and P.~Moln\'ar.
\newblock Social force model for pedestrian dynamics.
\newblock {\em Phys. Rev. E}, 51:4282--4286, May 1995.

\bibitem{huang2017densely}
G.~Huang, Z.~Liu, L.~Van Der~Maaten, and K.~Q. Weinberger.
\newblock Densely connected convolutional networks.
\newblock In {\em Proceedings of the IEEE conference on computer vision and
  pattern recognition}, pages 4700--4708, 2017.

\bibitem{huang2017body}
S.~Huang, X.~Li, Z.~Zhang, F.~Wu, S.~Gao, R.~Ji, and J.~Han.
\newblock Body structure aware deep crowd counting.
\newblock {\em IEEE Transactions on Image Processing}, 27(3):1049--1059, 2017.

\bibitem{jiang2019deepurbanevent}
R.~Jiang, X.~Song, D.~Huang, X.~Song, T.~Xia, Z.~Cai, Z.~Wang, K.-S. Kim, and
  R.~Shibasaki.
\newblock Deepurbanevent: A system for predicting citywide crowd dynamics at
  big events.
\newblock In {\em Proceedings of the 25th ACM SIGKDD International Conference
  on Knowledge Discovery \& Data Mining}, pages 2114--2122. ACM, 2019.

\bibitem{jin2018spatio}
W.~Jin, Y.~Lin, Z.~Wu, and H.~Wan.
\newblock Spatio-temporal recurrent convolutional networks for citywide
  short-term crowd flows prediction.
\newblock In {\em Proceedings of the 2nd International Conference on Compute
  and Data Analysis}, pages 28--35. ACM, 2018.

\bibitem{lecun1989backpropagation}
Y.~LeCun, B.~Boser, J.~S. Denker, D.~Henderson, R.~E. Howard, W.~Hubbard, and
  L.~D. Jackel.
\newblock Backpropagation applied to handwritten zip code recognition.
\newblock {\em Neural computation}, 1(4):541--551, 1989.

\bibitem{liu2018decidenet}
J.~Liu, C.~Gao, D.~Meng, and A.~G. Hauptmann.
\newblock Decidenet: Counting varying density crowds through attention guided
  detection and density estimation.
\newblock In {\em Proceedings of the IEEE Conference on Computer Vision and
  Pattern Recognition}, pages 5197--5206, 2018.

\bibitem{luo2018fast}
W.~Luo, B.~Yang, and R.~Urtasun.
\newblock Fast and furious: Real time end-to-end 3d detection, tracking and
  motion forecasting with a single convolutional net.
\newblock In {\em Proceedings of the IEEE conference on Computer Vision and
  Pattern Recognition}, pages 3569--3577, 2018.

\bibitem{rudenko2019human}
A.~Rudenko, L.~Palmieri, M.~Herman, K.~M. Kitani, D.~M. Gavrila, and K.~O.
  Arras.
\newblock Human motion trajectory prediction: A survey.
\newblock {\em arXiv preprint arXiv:1905.06113}, 2019.

\bibitem{DBLP:journals/corr/SindagiP17}
V.~Sindagi and V.~M. Patel.
\newblock A survey of recent advances in cnn-based single image crowd counting
  and density estimation.
\newblock {\em CoRR}, abs/1707.01202, 2017.

\bibitem{tripathi2019convolutional}
G.~Tripathi, K.~Singh, and D.~K. Vishwakarma.
\newblock Convolutional neural networks for crowd behaviour analysis: a survey.
\newblock {\em The Visual Computer}, 35(5):753--776, 2019.

\bibitem{WarpDriver}
D.~Wolinski, M.~C. Lin, and J.~Pettr{\'e}.
\newblock Warpdriver: Context-aware probabilistic motion prediction for crowd
  simulation.
\newblock {\em ACM Trans. Graph.}, 35(6):164:1--164:11, Nov. 2016.

\bibitem{yi2016pedestrian}
S.~Yi, H.~Li, and X.~Wang.
\newblock Pedestrian behavior understanding and prediction with deep neural
  networks.
\newblock In {\em European Conference on Computer Vision}, pages 263--279.
  Springer, 2016.

\bibitem{yogameena2017computer}
B.~Yogameena and C.~Nagananthini.
\newblock Computer vision based crowd disaster avoidance system: A survey.
\newblock {\em International journal of disaster risk reduction}, 22:95--129,
  2017.

\bibitem{zhang2019graphical}
J.~Zhang, K.~J. Shih, A.~Elgammal, A.~Tao, and B.~Catanzaro.
\newblock Graphical contrastive losses for scene graph parsing.
\newblock In {\em Proceedings of the IEEE Conference on Computer Vision and
  Pattern Recognition}, pages 11535--11543, 2019.

\bibitem{zhang2017deep}
J.~Zhang, Y.~Zheng, and D.~Qi.
\newblock Deep spatio-temporal residual networks for citywide crowd flows
  prediction.
\newblock In {\em Thirty-First AAAI Conference on Artificial Intelligence},
  2017.

\bibitem{PeriodicDensityPrediction}
A.~Zonoozi, J.~jae Kim, X.~li~Li, and G.~Cong.
\newblock Periodic-crn: A convolutional recurrent model for crowd density
  prediction with recurring periodic patterns.
\newblock In {\em IJCAI}, 2018.

\end{thebibliography}
